\documentclass{article}

    \PassOptionsToPackage{numbers, compress}{natbib}

    \usepackage[preprint]{neurips_2020}

\usepackage[utf8]{inputenc} 
\usepackage[T1]{fontenc}    
\usepackage{hyperref}       
\usepackage{url}            
\usepackage{booktabs}       
\usepackage{amsfonts}       
\usepackage{nicefrac}       
\usepackage{microtype}      

\usepackage{color}

\usepackage{multirow}
\usepackage{pifont}
\newcommand{\cmark}{\ding{51}}%
\newcommand{\xmark}{\ding{55}}%
\usepackage{times}
\usepackage{epsfig}
\usepackage{graphicx}
\usepackage{amsmath}
\usepackage{amssymb}
\usepackage{enumitem}
\usepackage{bm}
\usepackage{booktabs}
\usepackage{subcaption}
\usepackage{multirow}
\usepackage{sidecap}  

\newcommand{\etal}{\textit{et al}. }
\newcommand{\ie}{\textit{i}.\textit{e}.}

\title{Source Free Domain Adaptation with \\ Image Translation}

\author{%
  Yunzhong Hou, Liang Zheng\\
  Australian National University\\
  Australian Centre for Robotic Vision\\
  \texttt{\{firstname.lastname@anu.edu.au\}} \\
}

\begin{document}

\maketitle

\begin{abstract}
Effort in releasing large-scale datasets may be compromised by privacy and intellectual property considerations. A feasible alternative is to release pre-trained models instead. While these models are strong on their original task (source domain), their performance might degrade significantly when deployed directly in a new environment (target domain), which might not contain labels for training under realistic settings. Domain adaptation (DA) is a known solution to the domain gap problem, but usually requires labeled source data. 
In this paper, we study the problem of source free domain adaptation (SFDA), whose distinctive feature is that the source domain only provides a pre-trained model, but no source data. Being source free adds significant challenges to DA, especially when considering that the target dataset is unlabeled. 
To solve the SFDA problem, we propose an image translation approach that transfers the style of target images to that of unseen source images. To this end, we align the batch-wise feature statistics of generated images to that stored in batch normalization layers of the pre-trained model. Compared with directly classifying target images, higher accuracy is obtained with these style transferred images using the pre-trained model. On several image classification datasets, we show that the above-mentioned improvements are consistent and statistically significant. 
\end{abstract}

\section{Introduction}
\label{sec:intro}
Large-scale datasets are critical. However, oftentimes datasets are not ready for release due to privacy or intellectual property considerations, or just too large in size. In such cases, pre-trained models are the \emph{de-facto} choice of release. Pre-trained models not only serve as strong baselines for the original dataset, but also contain knowledge of the original dataset that enables many applications such as knowledge distillation~\cite{hinton2015distilling}, semi-supervised learning~\cite{erhan2010does}, and continual learning~\cite{parisi2019continual}. 

The distribution difference or domain gap between the training dataset (source domain) and the testing dataset (target domain) can degrade the performance of pre-trained models when directly applied. Oftentimes the target domain is unlabeled and prevents us from fine-tuning the model in a supervised manner. Domain adaptation (DA) can alleviate the influence of domain gap, where unsupervised DA specializes in unlabeled target domains. 
Many approaches are proposed for unsupervised DA~\cite{long2015learning,ganin2016domain,tzeng2017adversarial,bousmalis2017unsupervised,pmlr-v80-hoffman18a}. However, the majority of domain adaptation methods rely on a \emph{labeled source domain dataset}, which might not be available in many scenarios. 

\begin{figure*}
\centering
\includegraphics[width=0.9\linewidth]{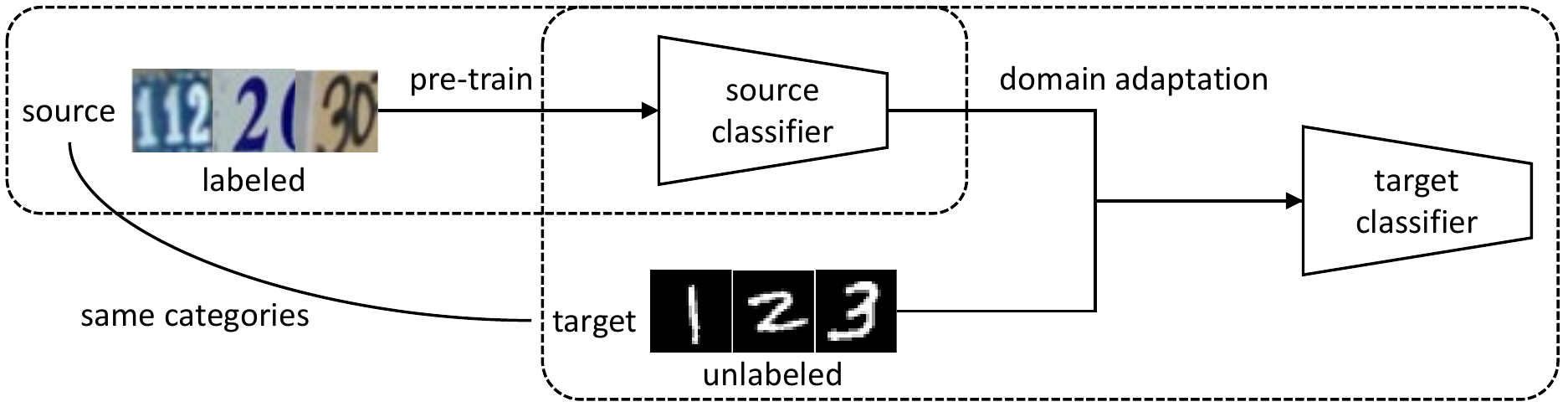}
\caption{
Problem setting for source free domain adaptation (SFDA). \textbf{Left:} Instead of source images, the system has access to only a pre-trained source classifier. \textbf{Right:} Then, the system adapts to an unlabeled target domain in the absence of source dataset. Both domains share the same categories.
}
\vspace{-3mm}
\label{fig:problem_setting}
\end{figure*}

On identifying these problems, we study on a demanding setting for domain adaptation, source free domain adaptation (SFDA). As shown in Fig.~\ref{fig:problem_setting}, aside from a pre-trained classifier on source domain, this setting allows no other cues (images and labels) from the source domain, and has an unlabeled target domain. We restrict the problem to a closed-set setting where the source and target domains share the same categories. 
Very few previous works have investigated the source free setting in DA. Chidlovskii \etal \cite{chidlovskii2016domain} use pseudo labels to fine-tune the source classifier. 
Li \etal \cite{li2018adaptive} change the statistics in batch normalization layers of the pre-trained classifier. Nelakurthi \etal  \cite{nelakurthi2018source} additionally use a few labeled target examples, which does not fall exactly under our assumption. Recently, Kundu \etal \cite{kundu2020universal} focus on universal domain adaptation (arbitrary relationship between source and target label sets) \cite{you2019universal}, but their approach needs a specifically designed module in the pre-trained classifier, which is a tough requirement for releasing pre-trained models.

In this paper, we propose a generative approach to SFDA: source free image translation. In a nutshell, instead of directly classifying the target images, we translate them into the \emph{unseen} source style before feeding them to the pre-trained source classifier. Similar to style transfer~\cite{huang2017arbitrary}, we generate source style images using feature map statistics. In the absence of source images and therefore their per-image statistics, we match the batch-wise feature statistics of style transferred images to that stored in the Batch Normalization (BN) layer of the source classifier instead. Compared with original target images, these style transferred images give higher accuracy on the source classifier. 

We demonstrate the effectiveness of our approach on digit datasets~\cite{lecun2010mnist,37648} and the VisDA dataset~\cite{visda2017}. Consistent and statistically significant improvements are achieved using the style transferred images. 

\section{Related Work}
\label{sec:related}

Domain adaptation (DA) methods aim to reduce the negative influence of domain gap between source and target domains. A recent research focus, unsupervised domain adaptation has a labeled source domain and an unlabeled target domain for training. 
Feature-level distribution alignment is a popular strategy for DA \cite{long2015learning,ganin2016domain,tzeng2017adversarial,saito2018maximum}. Long \etal \cite{long2015learning} use the maximum mean discrepancy (MMD) loss for this purpose. Ganin \etal \cite{ganin2016domain} propose the DANN method that has a gradient reversal layer and a discriminator. Tzeng \etal \cite{tzeng2017adversarial} propose an adversarial method, ADDA, with a loss function based on the generative adversarial network (GAN). 
Saito \etal \cite{saito2018maximum} align the task-specific decision boundaries of two classifiers. 
Pixel-level alignment with image generation is another effective choice in DA \cite{liu2016coupled,bousmalis2016domain,taigman2016unsupervised,shrivastava2017learning,bousmalis2017unsupervised,pmlr-v80-hoffman18a}. CoGAN \cite{liu2016coupled} learns the joint distribution of two domains by enforcing a weight-sharing constraint. 
DTN generates a source-styled image from a target image based on shared embedding and reconstruction loss \cite{taigman2016unsupervised}. Hoffman \etal propose the CyCADA \cite{pmlr-v80-hoffman18a} method by focusing on cross domain image generation and feature distribution alignment together. 
Other options are also investigated. For example, AdaBN \cite{li2018adaptive} simply adapts the statistics of the batch normalization layers to the target domain. French \etal \cite{french2018selfensembling} propose a self-ensemble method by mimicking the mean-teacher method \cite{tarvainen2017mean} for semi-supervised learning. Our work departs from unsupervised DA in that we do not require source domain images, only the source classifier instead.

Data-free knowledge distillation attracts much attention recently~\cite{lopes2017data,pmlr-v97-nayak19a,chen2019data,micaelli2019zero,haroush2019knowledge,yin2019dreaming}. Instead of distilling teacher knowledge on a given training dataset, data-free methods first \emph{generate} this training dataset and then learn on this generated dataset. Reconstructing training images can rely on aligning feature statistics \cite{lopes2017data,haroush2019knowledge,yin2019dreaming}, enforcing high teacher confidence~\cite{lopes2017data,pmlr-v97-nayak19a,chen2019data,haroush2019knowledge,yin2019dreaming}, and adversarial generation of hard examples for the student~\cite{micaelli2019zero,yin2019dreaming}. 
Specifically, in \cite{haroush2019knowledge,yin2019dreaming}, researchers match batch normalization statistics as additional regularization for the data-free image generation. Our work, while also assuming no access to source images, differs significantly from these works in that our image translation must not change the content, whereas data-free knowledge distillation just generates images without specifying the content and learns from the teacher decisions. 

Style transfer renders the same content in a different artistic style \cite{gatys2016image,johnson2016perceptual,luan2017deep,huang2017arbitrary,li2017universal}. In contrast to the GAN based image translation commonly adopted in pixel-level DA \cite{liu2016coupled,taigman2016unsupervised,pmlr-v80-hoffman18a}, style transfer usually features two losses: a content loss based on high-level feature maps, and a style loss based on feature map statistics of multiple intermediate layers. 
Huang \etal \cite{huang2017arbitrary} propose a real-time AdaIN method by changing the statistics in instance normalization layer for style transfer. 
The proposed method adopts the idea in AdaIN by using feature map mean and variance as style representation, but uses stored batch-wise statistics rather than unavailable per-image statistics.


\begin{table}[t]
\centering
\caption{
Problem setting comparison between SFDA and related fields of study. 
}
\small
\begin{tabular}{l|c|c|c}
\toprule
                                                                     & \multicolumn{1}{l|}{No source} & \multicolumn{1}{l|}{Unlabeled target} & \multicolumn{1}{l}{Closed-set} \\ \hline
Continual learning \cite{parisi2019continual}       & \cmark                  & -                         & -                               \\ \hline
Data-free knowledge distillation \cite{lopes2017data}                    & \cmark                  & -                         & \cmark           \\ \hline
Zero-shot DA \cite{peng2018zero}                    & \xmark                  & \xmark                         & \cmark           \\ \hline
Source free universal DA \cite{kundu2020universal} & \cmark                  & \cmark                         & \xmark           \\ \hline
Source free DA (SFDA)                                         & \cmark                  & \cmark                         & \cmark           \\
\bottomrule
\end{tabular}
\vspace{-3mm}
\label{tab:problem_setting}
\end{table}


\section{Method}
\label{sec:method}

To address source free domain adaptation, we propose source free image translation, a pixel-level DA method that transfers the style of target images to that of the unseen source images. 
In this section, we first compare the problem setting in SFDA with related fields of study in Section \ref{sec:sec:problem_setting}. Next, we give an overview of our approach in Section \ref{sec:sec:overview}. We then present our source free image translation approach in Section \ref{sec:sec:generator}. 
Finally, we provide further discussion in Section \ref{sec:sec:discussion}.

\subsection{Problem Setting}
\label{sec:sec:problem_setting}

We compare the problem setting of SFDA with several related fields of study in Table~\ref{tab:problem_setting}. 
In source free domain adaptation, we do not have access to source domain cues other than a pre-trained classifier. On top of that, we have an unlabeled target dataset, and assume the same categories (closed-set) between source and target domains. 
There are several fields of study related to SFDA. 
Continual learning~\cite{parisi2019continual} aims at not only learning continually from either same categories or new categories, but also not forgetting (not required in DA). 
Data-free knowledge distillation learns a new student model from scratch using teacher knowledge on the generated images~\cite{lopes2017data}, but does not deal with domain gaps. 
Zero-shot domain adaptation \cite{peng2018zero} has no access to target domain images during training, but can learn from task-irrelevant source-target domain pairs and task-relevant source domain (not source free). 
Source free \textit{universal} domain adaptation~\cite{kundu2020universal} focuses on the universal DA setting~\cite{you2019universal}. Rather than same source and target categories in closed-set setting, in universal DA, the relationship between the source labels and target labels can be arbitrary: closed-set, open-set, or no overlapping. 

\subsection{Overview}
\label{sec:sec:overview}
\begin{figure*}[t]
\centering
\includegraphics[width=0.9\linewidth]{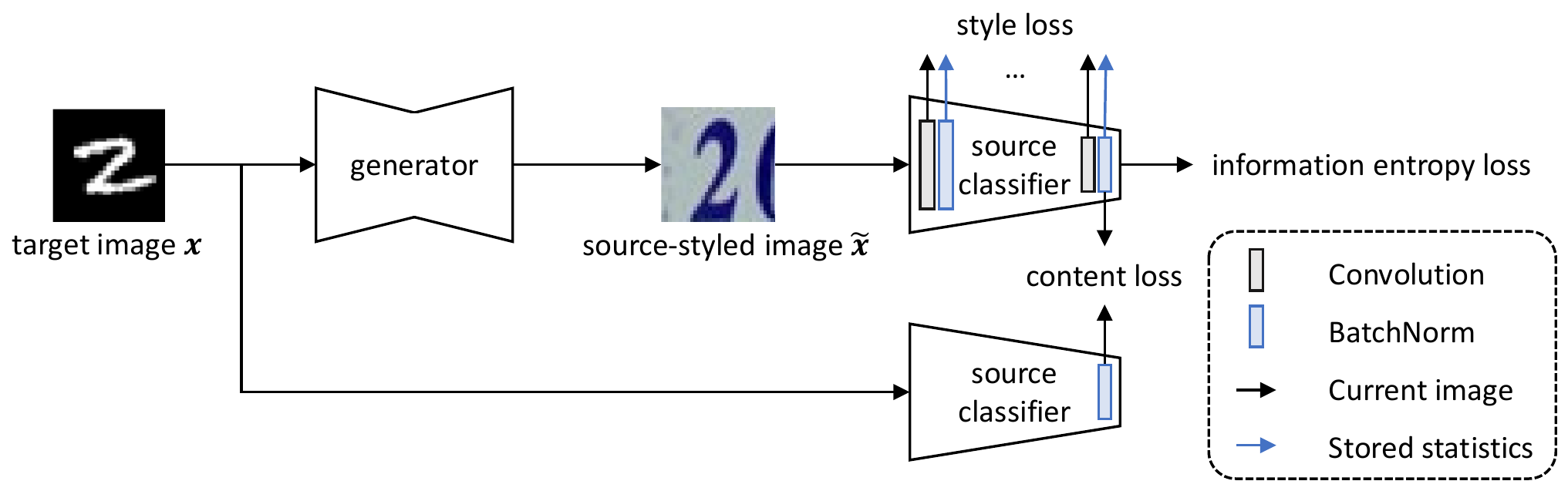}
\caption{
Network flow of source free image translation for SFDA. The \textbf{content loss} is computed as the difference of last-layer feature maps between target image $\bm x$ and source-styled image $\widetilde{\bm x}$. The \textbf{style loss} is computed as the average feature statistics difference of all layers. For layer $n\in \left\{1,2,...,N\right\}$, we compute difference between current batch of source-styled image feature map $f^n\left(\widetilde{\bm x}\right)$ statistics and the Batch Normalization layer statistics stored in the source classifier. 
The \textbf{information entropy loss} is based on the output probability vector $p\left(\widetilde{\bm x}\right)$.
}
\vspace{-3mm}
\label{fig:train_generator}
\end{figure*}

As shown in Fig.~\ref{fig:train_generator}, we train a generator $g\left(\cdot\right)$ to transfer the style of a target domain image $\bm x$ to the unseen source domain. During training, the generator is learned by aligning batch-wise feature statistics of style transferred images $\widetilde{\bm x} = g\left(\bm x\right)$ and that stored in Batch Normalization (BN) layers of the source classifier. During testing, the image translation works as a pre-process: instead of the original target image $\bm x$, the source-styled image $\widetilde{\bm x}$ is fed into the pre-trained source classifier $f\left(\cdot\right)$. We show in experiment that $\widetilde{\bm x}$ leads to higher accuracy than using the original target image $\bm x$. 

\subsection{Source Free Image Translation}
\label{sec:sec:generator}

In order to transfer target image to the unseen source style, we train the generator network $g\left(\cdot\right)$ in the following manner. For generator architecture design, we choose the residue architecture in CycleGAN~\cite{zhu2017unpaired}. During training (see Fig.~\ref{fig:train_generator}), we feed both the original target image $\bm x$ and the source-styled image $\widetilde{\bm x}$ into the pre-trained and fixed source classifier $f\left(\cdot\right)$. Given a style transferred image $\widetilde{\bm x}$, the source classifier outputs a feature map $f^n\left(\widetilde{\bm x}\right)$ for each layer $n\in \left\{1,2,...,N\right\}$, and a probability vector $p\left(\widetilde{\bm x}\right)$ for all $C$ classes. 
For loss design, we employ a content loss and a style loss on the feature maps $f^n\left(\widetilde{\bm x}\right)$ for style transfer, and an information entropy loss on the probability vector $p\left(\widetilde{\bm x}\right)$ to encourage higher source classifier confidence. The loss functions are described below.

\textbf{Content loss.}
Usually, in style transfer, the content loss encodes the difference of the feature maps in top layers between generated image and \textit{content} image. In our approach, we view the original target image $\bm x$ as \textit{content} image, and calculate the difference of last-layer features maps from the source classifier as the content loss,
\begin{equation}
    \mathcal{L}_\text{content} = \left\| f^N\left(\widetilde{\bm x}\right) - f^N\left(\bm x\right) \right\|_2,
\end{equation}
where $\left\|\cdot\right\|_2$ denotes the Euclidean distance. This loss ensures that the generated source style images have minimal change in its class label. 

\textbf{Style loss.} The style loss is usually computed as average statistics difference of feature maps, between the reconstructed image and the \textit{style} image \cite{gatys2016image,huang2017arbitrary}. The statistics can take the form of Gram matrix \cite{gatys2016image,johnson2016perceptual} or instance normalization (IN) mean and variance \cite{huang2017arbitrary}. Unlike the content loss that only considers high-level feature maps, 
the style loss considers multiple intermediate layers. 

In the context of domain adaptation, we refer to source images as \textit{style} images. However, in SFDA, the system has no access to source (style) images. 
Thus, we have to use a suitable form of statistics for style alignment. Batch normalization (BN) layers~\cite{10.5555/3045118.3045167} store the running mean and running variance for image batches, which are calculated on the feature maps of the training data. In light of this, rather than using IN statistics of an unavailable style image, we use BN statistics that are stored in pre-trained source classifier as a representative of the style of the source domain. We can then write the style loss for aligning batch-wise statistics of generated images with stored BN statistics as,
\begin{equation}
    \mathcal{L}_{\text{style}} = \frac{1}{N}\sum_{n=1}^{N}{\left\| \mu_\text{current}^n - \mu_\text{stored}^n \right\|_2 + \left\| \sigma_\text{current}^{n} - \sigma_\text{stored}^{n} \right\|_2},
\end{equation}
where at layer $n\in \left\{1,2,...,N\right\}$, $\mu_\text{current}^n$ and $\sigma_\text{current}^{n}$ denote the mean and standard variation of the current batch of source-styled images, respectively; $\mu_\text{stored}^n$ and $\sigma_\text{stored}^{n}$ denote the running mean and the square root of running variance stored in the source classifier, respectively. 

\textbf{Information entropy loss.}
Minimizing entropy \cite{grandvalet2005semi} of the label probability distribution proves to be effective in semi-supervised learning. In SFDA, we minimize this entropy loss to generate images that source classifier are more confident with (probability distribution more similar to one-hot), or in other words, more familiar with. For a style transferred image $\widetilde{\bm x}$, we calculate the information entropy loss on the probability distribution $p\left(\widetilde{\bm x}\right)$ estimated by the source classifier $f\left(\cdot\right)$ as,
\begin{equation}
    \mathcal{L}_\text{entropy} = -\sum_{c=1}^C{p_c\left(\widetilde{\bm x}\right)\log p_c\left(\widetilde{\bm x}\right)}, 
\end{equation}
where $p_c\left(\widetilde{\bm x}\right)$ denotes the probability from the source classifier for class $c\in\left\{1,2,...,C\right\}$. 

Combining the three aforementioned loss terms, we give the overall loss function for style transfer,
\begin{equation}
    \mathcal{L}_\text{generation} = \lambda_\text{content} \mathcal{L}_\text{content} + \lambda_\text{style} \mathcal{L}_\text{style} + \lambda_\text{entropy} \mathcal{L}_\text{entropy}, 
\end{equation}
where $\lambda_\text{content}$, $\lambda_\text{style}$, and $\lambda_\text{entropy}$ are scalar weights for balancing the three loss components.

\subsection{Discussion}
\label{sec:sec:discussion}

To reduce the domain shift, we transfer the style of the target images to that of the unseen source images without changing the content. 
To this end, we incorporate three loss terms: a content loss, a style loss, and an information entropy loss. Although mentioned in previous literature, they are changed according to the SFDA setting to serve different purposes, and are all necessary.

\textbf{The content loss ensures no semantics change.} In data-free knowledge distillation \cite{lopes2017data}, the student learns from teacher knowledge on generated images, and the image generation does not need to specify content or label \cite{micaelli2019zero,yin2019dreaming}. In contrast, for SFDA,
our method use style transfer as a pre-process before feeding the image to the source classifier during testing, which must not change the label of the original target image. Instead of ImageNet~\cite{imagenet_cvpr09} pre-trained VGG \cite{simonyan2014very} as in style transfer literature \cite{lopes2017data,huang2017arbitrary}, we use the pre-trained \textit{source classifier} to calculate the content loss, so as to ensure minimal label change during testing.

\textbf{The style loss with stored BN statistics guides the style transfer in SFDA.} 
Many pixel-level DA methods are based on GAN \cite{liu2016coupled,taigman2016unsupervised,pmlr-v80-hoffman18a}. However, the lack of source images makes it difficult to train a discriminator, further limiting the usage of GAN-based methods. Li \etal \cite{10.5555/3172077.3172198} draw a direct connection between style loss in style transfer and MMD in domain adaptation. As such, we use style loss to guide our pixel-level DA. In the absence of style images and their per-image statistics, we use the stored BN statistics in our style loss instead. 

\textbf{The information entropy loss enables the source classifier to work in a somewhat similar way to a discriminator.} For target to source image translation in a GAN-based pixel-level DA system  \cite{liu2016coupled,taigman2016unsupervised,pmlr-v80-hoffman18a}, the discriminator awards the generator for images whose styles are similar to the source (style) images, as these images should be \textit{easier} for the source classifier to recognize. In the source free setting where discriminators cannot be trained, the information entropy loss directly awards the generation of images that the source classifier can \textit{easily} recognize.  


In addition, we discuss potential limitations from our usage of stored batch-wise statistics, and compare with another SFDA method that also works with the stored BN statistics. 

\textbf{Limited randomness and diversity from stored batch-wise statistics.} 
Rather than using per-image statistics of random style images as in AdaIN \cite{huang2017arbitrary}, due to the lack of source (style) images, we use stored batch-wise feature statistics instead. 
In previous style transfer methods\cite{lopes2017data,huang2017arbitrary}, during training, style images are randomly sampled, giving different \textit{per-image} statistics and randomness in training. In contrast, the \textit{stored batch-wise} statistics do not change throughout the training, limiting the diversity and randomness, and causing difficulties in training. We will further examine the influence of this limited randomness in the experiments. 

\textbf{Comparison on different usage of stored BN statistics with AdaBN.}      
Both our style transfer approach and AdaBN~\cite{li2018adaptive} work with the BN running mean and variance stored in the source classifier, but are different in the way of using them in training and the network flow in testing. In training, we use the running mean and variance to guide style transfer, while AdaBN substitutes the stored source domain statistics with new target domain statistics to modify the target classifier. In testing, AdaBN feeds the original target images to the modified target classifier, whereas in our method, the source-styled images are fed into the source classifier directly.

\section{Experiment}
\label{sec:experiment}

\subsection{Experiment Setup}
\label{sec:sec:experiment_setup}

\textbf{Datasets.} We validate the proposed SFDA approach with four datasets. First, we evaluate SFDA for 10-class digits recognition on MNIST~\cite{lecun2010mnist}, USPS, and SVHN datasets~\cite{37648}. Then, we experiment on the 12-class VisDA dataset~\cite{visda2017}, which is the largest cross-domain object classification dataset to date. For digits DA MNIST $\rightarrow$ USPS, USPS $\rightarrow$ MNIST, and SVHN $\rightarrow$ MNIST, we use full training sets for learning, and evaluate on the test set in the target domain. For VisDA adaptation SYN $\rightarrow$ REAL, we use the synthetic dataset as source and the validation partition of the real-world dataset as target.

\textbf{Evaluation protocol.} We report the top-1 classification accuracy on the target domain (test set for the digits datasets; validation partition for the VisDA following previous works \cite{saito2018maximum,pinheiro2018unsupervised,kim2019unsupervised}).

\textbf{Implementation details.} For digits datasets, we choose a three-layer CNN classifier following \cite{long2018conditional}. All digits images are resized to $32\times32$ without train-time data augmentation, and fed into the system in a batch size of 128. For VisDA dataset, we choose the ResNet-50~\cite{7780459} architecture pre-trained on ImageNet~\cite{imagenet_cvpr09} as the classifier. All the VisDA images are resized and center cropped to $224\times224$, and we use a batch size of 32 with random crop and flip as data augmentation during training. 
For all the experiments, we use the following settings in training. We train the source classifier with SGD optimizer and a learning rate of $1\times10^{-2}$. The generator is trained with the Adam optimizer with a learning rate of $1\times10^{-4}$ for 30 epochs on digits datasets and 5 epochs on the VisDA dataset. We apply the cosine learning rate scheduling \cite{loshchilov2016sgdr} and set the batch size to 16 in VisDA specifically due to GPU memory considerations. 
In all our experiments, we set the loss weights as follows, $\lambda_\text{content}=1,\lambda_\text{style}=10,\lambda_\text{entropy}=0.1$. 
We finish all the experiments on an RTX-2080Ti GPU.

\begin{figure*}
\centering
    \begin{subfigure}[b]{\linewidth}
    \centering
        \includegraphics[width=\textwidth]{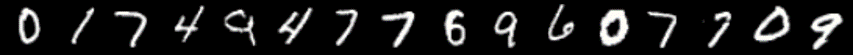}
        \caption{Original target images (MNIST)}
    \end{subfigure}
    
    \begin{subfigure}[b]{\linewidth}
    \centering
        \includegraphics[width=\textwidth]{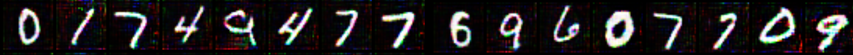}
        \caption{Generated source style images}
    \end{subfigure}
    
    \begin{subfigure}[b]{\linewidth}
    \centering
        \includegraphics[width=\textwidth]{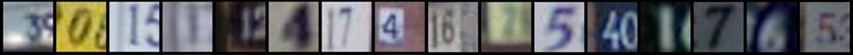}
        \caption{Unseen source images (SVHN)}
    \end{subfigure}
    
\caption{
Sample results of source free image translation on digits datasets SVHN $\rightarrow$ MNIST. 
For both Fig.~\ref{fig:generation_results_digits} and Fig.~\ref{fig:generation_results_visda}, we show in (a): original target images, (b): generated source style images, each of which corresponds to the original target image above it, and (c): the unseen source images. 
For gray-scale target images from MNIST, our style transfer adds random RGB colors to mimic the full-color style in the unseen source (SVHN) without changing the content. 
Note that the single image visual quality of the generated images is lower than previous methods that exploit both style images and content images in training. 
More discussions are available in Section~\ref{sec:sec:experiment_generator}.
}
\vspace{-3mm}
\label{fig:generation_results_digits}
\end{figure*}

\begin{figure*}
\centering
    \begin{subfigure}[b]{\linewidth}
    \centering
        \includegraphics[width=\textwidth]{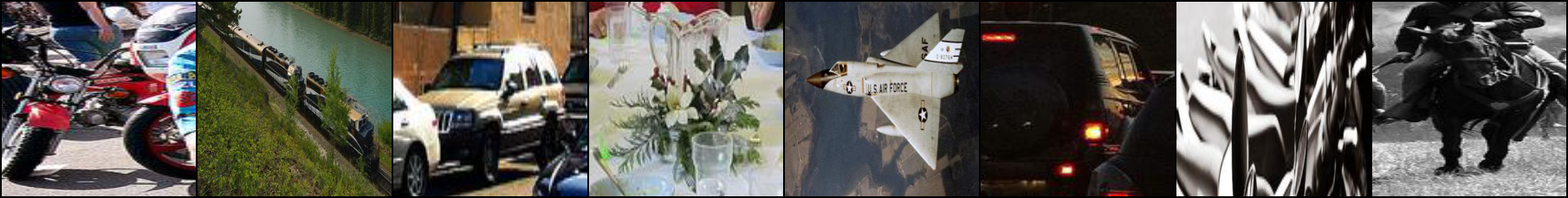}
        \caption{Original target images (real-world)}
    \end{subfigure}
    
    \begin{subfigure}[b]{\linewidth}
    \centering
        \includegraphics[width=\textwidth]{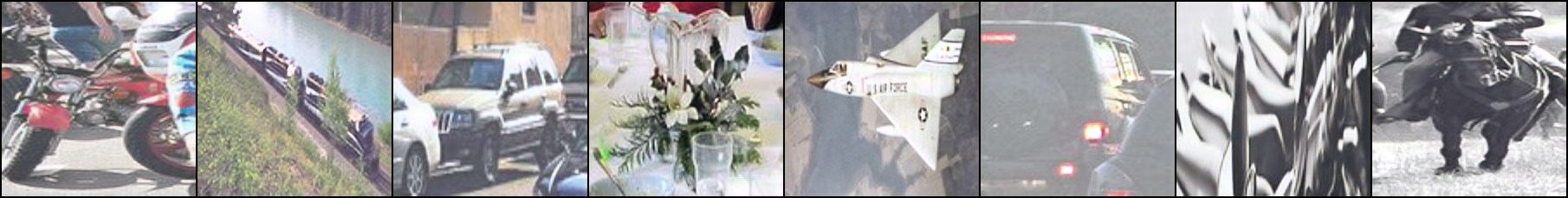}
        \caption{Generated source style images}
    \end{subfigure}
    
    \begin{subfigure}[b]{\linewidth}
    \centering
        \includegraphics[width=\textwidth]{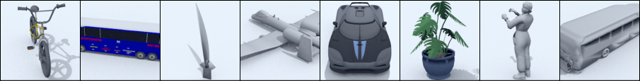}
        \caption{Unseen source images (synthetic)}
    \end{subfigure}
    
\caption{
Sample results of source free image translation on the VisDA dataset SYN $\rightarrow$ REAL. (a), (b), and (c) have the same meaning as in Fig. \ref{fig:generation_results_digits}. Our translation method increases the color saturation and brightness in target (real-world) to mimic the style of source (synthetic). 
}

\vspace{-2mm}
\label{fig:generation_results_visda}
\end{figure*}

\subsection{Evaluation}
\label{sec:sec:experiment_generator}

\textbf{Visualization of source free image translation results.}
We first visualize the style transferred images and compare with original target images and unseen source images. 
For digits datasets SVHN $\rightarrow$ MNIST (Fig.~\ref{fig:generation_results_digits}), 
the generator learns to add RGB colors to the gray-scale MNIST (target) images to mimic the full-color SVHN (source) images. 
For VisDA dataset SYN $\rightarrow$ REAL (Fig.~\ref{fig:generation_results_visda}), the generator learns to change the color saturation and brightness of the real-world (target) images so that they are more similar to the highly-saturated colors in the well-lit synthetic (source) scenario. 
These image generation results also show that the generator brings minimal content change to the original target images, which enables us to use the style transfer as a pre-process for target images in testing. 

In terms of visual quality of single images, our method is inferior to existing generation methods that require images from both domains \cite{gatys2016image,huang2017arbitrary,pmlr-v80-hoffman18a,taigman2016unsupervised}. 
As mentioned in Section~\ref{sec:sec:generator}, due to the lack of source images, we are unable to either adopt GAN-based methods with discriminator loss for each image, or take the style transfer approach with per-image statistics as the style loss, both of which provides \textit{per-image} supervision. Instead, our approach generates source style images with \textit{batch-wise} supervision from feature maps statistics and the stored BN running mean and variance. 
Batch-wise supervision cannot regularize single images independently. 
Given the lack of per-image supervision in the proposed method, its single image visual quality is thus limited. 
Nonetheless, we feed these style transferred image to the source classifier for evaluation during testing, and single image visual quality is not our top priority.

\begin{table}[t]
\centering
\caption{Comparisons of various DA methods and SFDA methods. ``No DA'' means directly feeding original target images to the source classifier. ``Ours'' indicates feeding style transferred images to the source classifier. $\star\star\star$ means
that the accuracy difference between ``No DA'' and our method is statistically very significant (\ie, p-value < 0.001). $\dag$ means the results are from our implementation. 
}
\resizebox{\linewidth}{!}{
\begin{tabular}{l|c|c|c|c|c}
\toprule
\multirow{2}{*}{Methods}                                        & \multicolumn{1}{l|}{\multirow{2}{*}{No source}} & \multicolumn{3}{c|}{Digits}                                                                                                                   & \multicolumn{1}{c}{VisDA}                  \\ \cline{3-6} 
                                                                & \multicolumn{1}{l|}{}                           & \multicolumn{1}{l|}{MNIST $\rightarrow$ USPS} & \multicolumn{1}{l|}{USPS $\rightarrow$ MNIST} & \multicolumn{1}{l|}{SVHN $\rightarrow$ MNIST} & \multicolumn{1}{l}{SYN $\rightarrow$ REAL} \\ \hline
MMD$^\dag$ \cite{long2015learning}         & \xmark               & 81.8                            & 76.0                              & 74.9                            & 58.1                          \\ \hline
DANN$^\dag$ \cite{ganin2016domain}         & \xmark              & 88.6                            & 89.6                            & 84.7                            & 60.5                          \\ \hline
CoGAN \cite{liu2016coupled}   & \xmark     & 91.2                            & 89.1                            & -                            & -                          \\ \hline
CyCADA (pixel only) \cite{pmlr-v80-hoffman18a}   & \xmark     & \textbf{95.6}                            & \textbf{96.4}                            & 70.3                            & -                          \\ \hline\hline
Fine-tuning$^\dag$ \cite{chidlovskii2016domain}     & \cmark          & 83.7                            & 70.2                            & 80.6                            & 55.1                          \\ \hline
AdaBN$^\dag$ \cite{li2018adaptive}        & \cmark              & 77.0                              & 67.8                            & 71.9                            & 56.7                          \\ \hline\hline
No DA (baseline)        & \cmark              & 80.4                            & 63.1                            & 73.3                            & 52.2                          \\ \hline
Ours         & \cmark         & 81.9 ($\star\star\star$)                            & 68.9                            & 84.6                            & 54.0 ($\star\star\star$)                           \\ \hline
Ours + fine-tuning  & \cmark & 84.7                            & 82.3                           & \textbf{90.4}                            & \textbf{63.5}                          \\ 
\bottomrule
\end{tabular}
}
\label{tab:results}
\vspace{-2mm}
\end{table}

\textbf{Improvements over the baseline.}
We compare the proposed method with the baseline. In the baseline, no domain adaptation is conducted, and we directly classify the target images using the source classifier. 
Results are presented in Table~\ref{tab:results}. Under all four DA scenarios,  
when feeding our style transferred image to the source classifier (``Ours'') instead of original target images (``No DA''), we witness consistent improvements in classification accuracy over the baseline. Specifically, the increase is +1.5\%, +5.8\%, +11.3\%, and +1.8\% on MNIST $\rightarrow$ USPS, USPS $\rightarrow$ MNIST, SVHN $\rightarrow$ MNIST, and SYN $\rightarrow$ REAL, respectively. 
For MNIST $\rightarrow$ USPS and SYN $\rightarrow$ REAL, while the improvement (+1.5\% and +1.8\%) is smaller, we find them statistically very significant (\ie, p-value < 0.001) through repeating the experiment 5 times.

\emph{Discussions.}
As mentioned in Section~\ref{sec:sec:discussion}, in previous style transfer methods \cite{lopes2017data,huang2017arbitrary}, per-image statistics from random style images provide randomness and diversity during training.
On the contrary, in the proposed method, \textit{stored} BN statistics do not change during training, limiting the randomness and diversity, and causes difficulties in generator training. 
Given the large domain gap in SYN $\rightarrow$ REAL and small-scale target dataset (USPS) in MNIST $\rightarrow$ USPS, this is particularly pronounced. 
For SYN $\rightarrow$ REAL, the domain gaps are relatively large and the images in two domains share minimal visual similarities. Given the large domain gap, the difficulties in training with only limited diversity from stored BN statistics are more pronounced in the small performance increase. 
For MNIST $\rightarrow$ USPS, the target domain USPS is a small-scale dataset with only 7291 training images. Without randomness in style image statistics (we use stored BN statistics instead) during training, diversity and randomness can only come from the target dataset. 
As such, limited diversity in the small-scale target domain adds even more difficulties in training. 

\textbf{Comparison with image translation methods that requires source images.}
We compare the classification accuracy obtained by our image translation method with other pixel-level DA methods such as CoGAN \cite{liu2016coupled} and CyCADA (pixel only) \cite{pmlr-v80-hoffman18a} that require source domain images. Results are summarized in Table \ref{tab:results}. We have two observations. 

First, the classification accuracy of our method is generally inferior to CoGAN and CyCADA, such as on MNIST $\rightarrow$ USPS and USPS $\rightarrow$ MNIST. This is expected, because their methods exploit the source images while ours does not. Other than the foreground-background ratio difference, MNIST and USPS are very similar visually, which leads to similar BN statistics between source and target and increases difficulties for our \textit{style loss}. Theoretically, the performance of CoGAN and CyCADA should be upper-bound of our source free method. 
Second, interestingly, we find that our method yields higher accuracy than CyCADA under SVHN $\rightarrow$ MNIST. 
Unlike MNIST and USPS that have similar BN statistics, the full-color SVHN and gray-scale MNIST have very different feature statistics, allowing our batch-wise \textit{style loss} to its full potential. 
In addition, CycleGAN generates images to fool the \textit{discriminator}, not to directly make them easier for the \textit{source classifier} to recognize. The two concepts might not be equivalent since the discriminator and the source classifier might focus on different aspects of images. In contrast, our method performs better with credit to the information entropy loss that directly encourages generating images the \textit{source classifier} is more confident with. 

\textbf{Ablation study.} 
We examine the necessity of the three loss functions using the VisDA dataset. The results are shown in Table 3. First, the removal of content loss leads to complete failure. Based on the assumption that the generator does not change the content, we feed the style transferred image to the source classifier during testing. Without this content consistency, this test-time pre-process becomes meaningless. Second, removing the style loss leads to performance drop by -0.4\%, as this removes all guidance from the unseen source dataset. Third, the removal of the entropy loss leads to a -0.3\% performance drop, which indicates its effectiveness. 
Though the performance difference from the exclusion of style loss and information entropy loss are small (-0.4\% and -0.3\%), we find them statistically significant (\ie, p-value<0.05) over 5 runs.

\textbf{Comparison with existing SFDA methods.} 
We re-implemented SFDA methods pseudo label fine-tuning~\cite{chidlovskii2016domain} and AdaBN~\cite{li2018adaptive}. These two methods amend the classifier, while ours does not, and our method achieves competitive performance. Note that our method cannot be combined with AdaBN. It is because AdaBN modifies source classifier to match the target statistics, while our strategy is to translate target images to the source style. That is, the direction of adaptation between AdaBN and our method is opposite. Below, we briefly show the effect of combining our method with fine-tuning~\cite{chidlovskii2016domain}.

\begin{table}[t]
\centering
\caption{
Ablation study for source free image translation. $\star$ means the accuracy difference between ``Ours'' (using all three loss terms) and the variants are statistically significant (\ie, p-value<0.05).
}
\small
\begin{tabular}{l|l|c|c|c|c|c}
\toprule
\multirow{3}{*}{Configuration} & Content loss & \xmark & \xmark & \cmark & \cmark & \cmark \\ \cline{2-7} 
                               & Style loss   & \xmark & \cmark & \xmark & \cmark & \cmark \\ \cline{2-7} 
                               & Entropy loss & \xmark & \cmark & \cmark & \xmark & \cmark \\ \hline
Performance                    & SYN $\rightarrow$ REAL     & 52.2  & 8.2      & 53.6 ($\star$)  & 53.7 ($\star$) & 54.0  \\
\bottomrule
\end{tabular}
\label{tab:ablation}
\vspace{-2mm}
\end{table}

\textbf{Further usage of source-styled images.}
We explore the possibility of mining pseudo labels from the generated images and fine-tuning the classifier. We follow fine-tuning method based on pseudo labels in \cite{chidlovskii2016domain}. Specifically, to calculate pseudo labels, in addition to the restriction of confidence being higher than 0.95, we only consider those target image whose labels agree with their corresponding source image label. 
As shown in Table~\ref{tab:results}, under the SFDA setting, the fine-tuned target classifier (``Ours + fine-tuning'') outperforms our implementation of \cite{chidlovskii2016domain}, which only uses pseudo labels from the target images for fine-tuning. 
This proves the inclusion of a second cue brings consistent improvements. 
We also note that, when pseudo labels are used, our system is even comparable to two feature-level DA methods MMD~\cite{long2015learning} and DANN~\cite{ganin2016domain} which requires a labeled source dataset.

\section{Conclusion}

In this paper, we study a problem that is of practical and research significance but rarely investigated: source free domain adaptation (SFDA). SFDA aims to adapt source domain knowledge to an unlabeled target domain in the absence of source data. Rather than fine-tuning pre-trained classifier with pseudo labels, or changing the BN statistics in the classifier as done in existing SFDA methods, we propose a source free image translation approach. We use BN statistics stored in the source classifier to represent and infer the style of the unseen source domain, and generate images whose batch-wise feature statistics match those BN statistics. The style transferred images can be directly fed into the pre-trained classifier to improve recognition accuracy. Our approach shows consistent and statistically significant improvement on several datasets and encourages further study for this problem.

\section*{Broader Impact}

Large scale datasets are essential to neural networks. However, privacy and intellectual property concerns oftentimes compromise the effort in releasing such datasets. In this paper, we study the problem of source free domain adaptation: bridging the domain gap to the unlabeled target domain, in the absence of source domain data. The proposed method enables us to effectively extract knowledge from the pre-trained model, thus lessening the urge to release labeled source domain datasets. 

\textbf{a) who may benefit from this research}: The idea of source free domain adaptation helps to protect the privacy of the people in certain datasets (from being released), and the intellectual property of the organization collecting the dataset (from releasing the dataset).

\textbf{b) who may be put at disadvantage from this research}: The authors cannot think of people or organizations whose interests might be affected as a direct outcome of this research. 

\textbf{c) what are the consequences of failure of the system}: The failure of the system has no severe consequence as the user can always retreat to feeding target images with the pre-trained source classifier, which means no performance improvements in the worst scenario. 

\textbf{d) whether the task/method leverages biases in the data}: The proposed method does not leverage biases in the dataset.

\bibliographystyle{unsrt}
\bibliography{refs}

\end{document}